\documentclass[10pt,twocolumn,letterpaper]{article}

\usepackage{cvpr}      

\usepackage[dvipsnames]{xcolor}

\definecolor{cvprblue}{rgb}{0.21,0.49,0.74}
\usepackage[pagebackref,breaklinks,colorlinks,citecolor=cvprblue]{hyperref}

\def\paperID{1872} 
\def\confName{CVPR}
\def\confYear{2024}

\title{Multimodal Large Language Model for Visual Navigation}

\author{
Yao-Hung Hubert Tsai$^\dagger$, Vansh Dhar$^\dagger$, Hugues Thomas$^\dagger$, Jialu Li$^\ddagger$\thanks{Work done during internship at Apple.}\,\,\,, Bowen Zhang$^\dagger$, Jian Zhang$^\dagger$ \\
  $^\dagger$Apple, $^\ddagger$University of North Carolina, Chapel Hill \\ {\small \tt  \{yaohung\_tsai,v\_dhar,hthomas23,bowen\_zhang4,jianz\}@apple.com,jialuli@cs.unc.edu} 
}

\begin{document}
\maketitle

\begin{abstract}
    Recent efforts to enable visual navigation using large language models have mainly focused on developing complex prompt systems. These systems incorporate instructions, observations, and history into massive text prompts, which are then combined with pre-trained large language models to facilitate visual navigation. In contrast, our approach aims to fine-tune large language models for visual navigation without extensive prompt engineering. Our design involves a simple text prompt, current observations, and a history collector model that gathers information from previous observations as input. For output, our design provides a probability distribution of possible actions that the agent can take during navigation. We train our model using human demonstrations and collision signals from the Habitat-Matterport 3D Dataset (HM3D). Experimental results demonstrate that our method outperforms state-of-the-art behavior cloning methods and effectively reduces collision rates.
\end{abstract}

\section{Introduction}

Visual navigation is a crucial feature for mobile agents, allowing them to process visual inputs and generate corresponding actions~\cite{bonin2008visual}. This technology finds applications in various fields, including elder care~\cite{ribeiro2021charmie}, autonomous driving~\cite{yurtsever2020survey}, and logistics delivery~\cite{chen2021adoption}. However, solving visual navigation is a complex task that requires a comprehensive understanding of different environments and the implementation of safety measures to protect both the agent and the surrounding objects~\cite{montello2005navigation}.

In recent years, the emergence of large language models (LLMs) has transformed artificial intelligence and business~\cite{devlin2018bert}. These models have found applications in document drafting~\cite{alafnan2023chatgpt}, storytelling~\cite{thorp2023chatgpt}, grammar checking~\cite{wu2023chatgpt}, and more. Researchers have also explored the use of LLMs for visual navigation, focusing on developing complex prompt systems~\cite{shah2023lm,huang2023visual,yu2023l3mvn,zhou2023navgpt}. These systems incorporate instructions, observations, and history into text prompts, which are then combined with pre-trained LLMs to facilitate visual navigation. However, a limitation of this approach is that pre-trained LLMs are typically trained only with text data and may not be best suited for tasks that require an understanding of other modalities~\cite{wu2023visual}, such as visual observations, GPS information, and compass data.

To address this limitation, recent work has focused on fine-tuning LLMs using additional image-text pairs~\cite{zhu2023minigpt,li2023blip,dai2023instructblip,liu2023visual}. This approach enables LLMs to answer questions about images~\cite{zhu2023minigpt,li2023blip,dai2023instructblip} or generate stories that interleave text and images~\cite{liu2023visual}. Building upon this, we propose to fine-tune LLMs specifically for visual navigation using observation-action pairs. During inference, LLMs directly process observations and generate low-level guidelines for the agent to follow, eliminating the need for extensive prompt system design.

\begin{figure}[t!]
    \centering
    \includegraphics[width=0.95\linewidth]{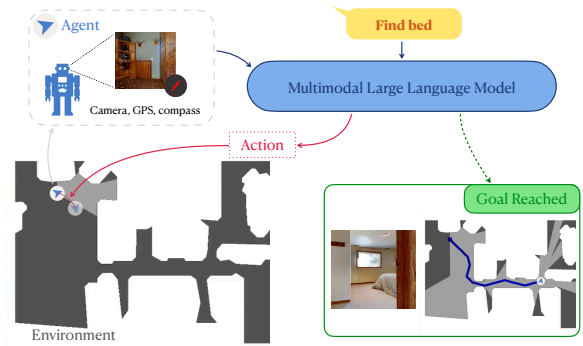}
    \vspace{-2mm}
    \caption{Our approach leverages a finetune multimodal large language model to solve object goal navigation.}
    \vspace{-4mm}
    \label{fig:arch}
\end{figure}

Our approach involves utilizing a simple text prompt, current observations (including visual inputs, GPS, and compass values), and a history collector model that gathers information from previous observations. These inputs are transformed into prompt tokens, current observation tokens, and history tokens. The large language model then processes these tokens and outputs a probability distribution of possible actions for the agent during navigation. For training, we use human demonstrations on the Habitat-Matterport 3D Dataset (HM3D)~\cite{ramakrishnan2021habitat} to form the probability of actions based on 1) human-demonstrated actions, 2) action probability distributions from state-of-the-art behavior cloning methods, and 3) collision signals.

In our experiments, we compare our approach with state-of-the-art behavior cloning methods and observe significant improvements in object goal navigation. We also find that having the large language model output a probability distribution over actions leads to better performance compared to directly outputting the action itself. Additionally, by considering collision signals during training, we observe a decrease in the number of collisions during visual navigation.
\section{Related Work}

This paper covers a wide range of literature. Below, we will discuss them in various topics.

{\bf Visual Navigation.} There are three important components in visual navigation: map building, localization, and path planning~\cite{bonin2008visual}. Map building involves the agent creating a map of the environment, localization involves the agent determining its position on the map, and path planning involves the agent deciding its actions based on the current context. In some scenarios where a pre-built map already exists, approaches like RTAB-Map \cite{labbe2019rtab} perform localization and path planning. However, in most real-world scenarios, maps are not provided, and SLAM systems~\cite{pritsker1984introduction} offer a solution by simultaneously building maps and performing localization. While classic approaches like orb-SLAM~\cite{mur2017orb} or LSD-SLAM~\cite{engel2014lsd} perform well, there is a growing trend of incorporating differentiable models, such as deep neural networks, into SLAM systems~\cite{chaplot2020learning,chaplot2020neural}. Furthermore, recent work has demonstrated that explicit map building and path planning are not necessary, and directly training reactive policies using recurrent neural networks like GRU~\cite{chung2014empirical} can achieve excellent performance~\cite{chaplot2020semantic,ramrakhya2023pirlnav}. Our method is similar to these approaches, but we use LLMs to train a reactive policy. During inference, our method outputs a probability distribution for the actions, and we select the action with the highest probability.

{\bf Large Language Models for Visual Navigation.} There have been several studies on visual navigation using LLMs. LM-Nav~\cite{shah2023lm} utilizes LLMs to extract landmarks from free-form navigation instructions. These landmarks are then passed to a vision-and-language model for grounding and a visual navigation model for navigation planning. L3MVN~\cite{yu2023l3mvn} proposes a method that calculates the entropy of objects in each frontier using a semantic segmentation model. This entropy is represented as query strings, and LLMs are used to determine a more relevant frontier. NavGPT~\cite{zhou2023navgpt} and another recent approach~\cite{vemprala2023chatgpt} interact with different visual foundation models to handle multimodal inputs. They also incorporate a history buffer and an LLM summarizer to handle the history, and aggregate information from various sources through a prompt manager. However, these approaches heavily rely on prompt engineering for LLMs and do not fine-tune the LLMs. In contrast, our method does not require extensive prompt engineering and directly fine-tunes LLMs for visual navigation policy.

{\bf Multimodal Large Language Models.} Performing visual navigation using LLMs requires LLMs to understand modalities beyond text. In this context, we discuss approaches that involve fine-tuning LLMs with image-text pairs to enhance their visual capabilities. MiniGPT4~\cite{zhu2023minigpt} proposes fine-tuning the pre-trained Llama~\cite{touvron2023llama} model using curated image-text pairs. It utilizes the visual encoder and q-former from BLIP2~\cite{li2023blip}, adds a trainable linear layer to transform visual features into visual tokens, inserts the visual tokens and text tokens from the text prompt into Llama~\cite{touvron2023llama}, and conducts the training. InstructBlip~\cite{dai2023instructblip} extends the idea of MiniGPT4 by training LLMs with high-quality image-text pairs. InstructBlip collects 26 publicly available datasets covering various tasks and capabilities and converts them into an instruction tuning format for fine-tuning LLMs. Similar to MiniGPT4 and InstructBlip, our method involves creating pairs between agent observations and actions, which we use to fine-tune LLMs. We consider observations from the visual image, compass values, and GPS information. Text is used to represent actions, such as ``go forward'' or ``turn right''.

{\bf Large Language Models for Robotics.} In this discussion, we explore the use of large language models (LLMs) for general robotics control. Palm-e~\cite{driess2023palm} proposes inputting tokens from various modalities (such as images, neural 3D representations, or states), along with text tokens, into LLMs. The model then generates high-level robotics instructions for tasks such as mobile manipulation, task and motion planning, and tabletop manipulation. In contrast, Instruct2Act~\cite{huang2023instruct2act} generates Python programs that form a complete perception, planning, and action loop for robotic tasks. Moving further, RT-2~\cite{brohan2023rt} generates low-level actions for robots, enabling closed-loop control. While this paper does not solve general-purpose robotics tasks, it focuses on visual navigation, which requires exploration in unseen environments, unlike the tasks studied in these works. It is worth noting that our method aligns with the approach of RT-2, as we generate low-level actions (in the form of a probability distribution) for the robot to execute. 

\begin{figure*}[t!]
    \centering
    \includegraphics[width=0.8\linewidth]{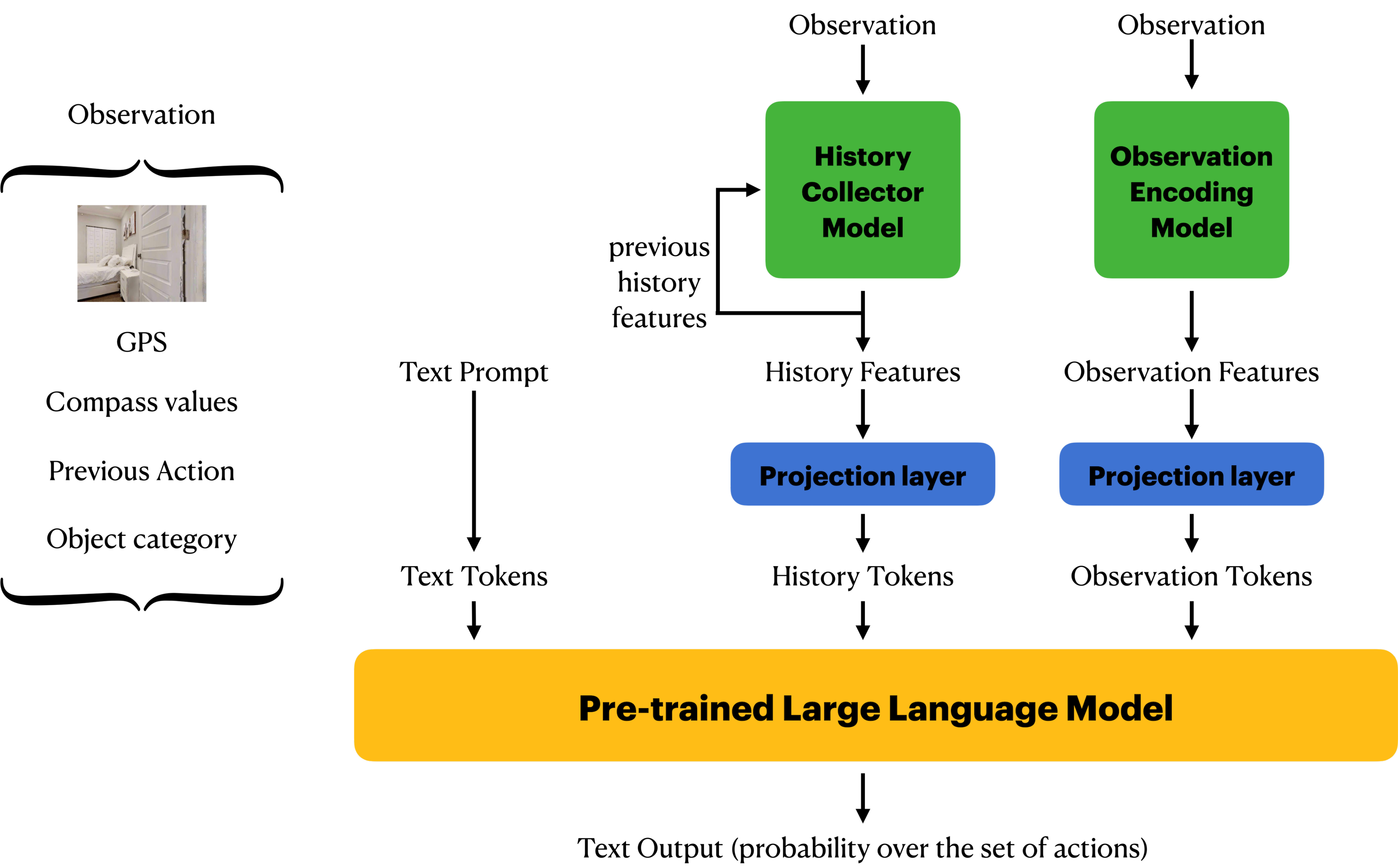}
    \vspace{-2mm}
    \caption{Architecture for fine-tuning large language models for visual navigation. The history collector model is responsible for encoding history features from the current observation and past history. The observation encoding model encodes observation features. The projection layer transforms history tokens and observation tokens from history and observation features, respectively. The text prompt is used to provide hints to the large language models (LLMs) for visual navigation. The pre-trained large language model takes text tokens, history tokens, and observation tokens as input, and generates a probability distribution over a set of actions as text output.}
    \label{fig:arch}
\end{figure*}

\section{Proposed Method}

Our method involves fine-tuning Large Language Models (LLMs) using pairs of observations and actions from a visual navigation agent. Our proposed architecture does the following: firstly, we have an observation encoding model that converts observations into observation tokens; secondly, we have a history collector model that gathers past observations as history and transforms this information into history tokens. Lastly, we have a pre-trained large language model that takes in text tokens from text prompts, observation tokens, and history tokens. It then outputs a probability distribution over the set of actions. During training, we utilize a human demonstration dataset. We construct the probability distribution using labels from the dataset, action outputs from a state-of-the-art behavior cloning method, and collision signals.

\subsection{Dataset}

Our task is object goal navigation, with object categories including ``chair'', ``bed'', ``plant'', ``toilet'', ``tv\_monitor'', and ``sofa''. In simple terms, the agent is required to navigate in an environment based on a given object goal. If the agent successfully reaches any objects within a distance of 1 meter that belong to the specified object category, it is considered a success.  

For fine-tuning, we use the human demonstration dataset curated by a recent work~\cite{ramrakhya2023pirlnav} using environments from the Habitat-Matterport 3D Research Dataset (HM3D)~\cite{ramakrishnan2021habitat} within the Habitat-sim simulator~\cite{habitat19iccv}. The dataset contains $77$k human demonstrations from $80$ training scenes. 

In the Habitat-sim simulator~\cite{habitat19iccv,szot2021habitat}, we consider the following observations: RGB visual image, compass values, and GPS values. The available actions are: ``stop'', ``go forward by 25 centimeters'', ``turn right by 30 degrees'', ``turn left by 30 degrees'', ``look up by 30 degrees'', and ``look down by 30 degrees''. The simulator also provides collision information. It is important to note that when a collision occurs, the agent remains at the same location in the simulator.

\subsection{Architecture}

We present the architecture in Figure~\ref{fig:arch}. The architecture consists of five core modules: the history collector model, the observation encoding model, the projection layers, the text prompt, and the pre-trained large language model. The history collector model is responsible for encoding historical features from the current observation and past history. The observation encoding model encodes observation features. The projection layer transforms history tokens and observation tokens from history and observation features, respectively. The text prompt is used to provide hints to the large language models (LLMs) for visual navigation. The prompt is transformed into text tokens using a text tokenizer. The pre-trained large language model takes text tokens, history tokens, and observation tokens as input and generates a probability distribution over a set of actions as text output.

\subsubsection{History Collector Model}
\label{subsubsec:history_collector}

The history collector model generates history features that carry meaningful information from the beginning of a training episode to the current time step. After evaluating different model combinations, we use ResNet-50~\cite{he2016deep} to encode visual features. Linear layers are used to encode GPS, compass values, previous action, and object category into corresponding features. These features are then transformed into history features using GRUs~\cite{chung2014empirical}.

We pretrain the history collector model using the behavior cloning method described in the paper~\cite{ramrakhya2023pirlnav} on the human demonstration dataset. It is important to note that the history collector model can be seen as a standalone model that generates actions from observations. We will provide further details on its performance in the experimental section. During the fine-tuning of the large language model with visual navigation, the weights of this history collector model remain fixed.

\subsubsection{Observation Encoding Model}

We utilize the pre-trained ViT~\cite{dosovitskiy2020image} and the Q-former~\cite{li2023blip} from the BLIP-2 model~\cite{li2023blip} as our observation encoding model for encoding visual images into our observation features. It is important to note that we have not observed any benefits in including information from GPS, compass values, or previous actions in our observation features. We believe this is because GPS, compass values, and previous actions are only meaningful when considered as a sequence. However, since our observation encoding model is designed to process only the current observation, we have chosen to include only the current visual image. We keep the weights of the observation encoding model fixed when fine-tuning the large language model for visual navigation.

\subsubsection{Projection Layers}

During the fine-tuning process, only the projection layers are trained. We use the Q-former~\cite{li2023blip} followed by a linear layer to project history features into 32 history tokens. Additionally, we use a linear layer to project observation features into 32 observation tokens.

\subsubsection{Text Prompt}

We provide a list of text prompts which are paraphrased with each other (using ChatGPT for paraphrasing). An example of the text prompt is:

{\it ``Imagine you are a robot, and you are navigating to find $\langle$ Goal $\rangle$ $\langle$ GoalHere $\rangle$ $\langle$ $\slash$ Goal $\rangle$ . With current observation $\langle$ Img $\rangle$ $\langle$ ImageHere $\rangle$ $\langle$ $\slash$ Img $\rangle$ , history tokens $\langle$ History $\rangle$ $\langle$ HistoryHere $\rangle$ $\langle$ $\slash$ History $\rangle$ , and suggested actions probabilities $\langle$ ActionProb $\rangle$ $\langle$ ActionProbHere $\rangle$ $\langle$ $\slash$ ActionProb $\rangle$ , please plan out your following action.''}

In this text prompt, {\it $\langle$ GoalHere $\rangle$} represents the object category. {\it $\langle$ ImageHere $\rangle$} represents the observation tokens. {\it $\langle$ HistoryHere $\rangle$} represents the history tokens. {\it $\langle$ ActionProbHere $\rangle$} represents the action probability output from the history collector model (See Section \ref{subsubsec:history_collector}). We present it via text, and an example for it is 

{\it ``Stop with probability 0.03, move forward with probability 0.44, turn left with probability 0.28, turn right with probability 0.21, look up with probability 0.03, and look down with probability 0.01''}

\subsubsection{Pre-trained Large Language Model}

We consider the pre-trained LLama-13B model~\cite{meta2023introducing} as a large language model that has its weights fixed during the fine-tuning process.

\subsection{Fine-tuning Paradigm}
\label{subsec:finetune}

We perform $80$k iterations for fine-tuning, with each iteration considering a batch size of $6$ observation-action pairs. Each episode in the human demonstration contains around $50$ to $100$ time steps. Therefore, the fine-tuning is conducted over $4.8$k to $9.6$k episodes out of the total of $77$k episodes in our dataset. For the output, we perform the following steps to construct the probability over the set of actions. 

Firstly, we use the state-of-the-art (SOTA) behavior cloning method from the paper~\cite{ramrakhya2023pirlnav} to compute the output probability over the actions, starting from the beginning of an episode until the current observation. We denote this probability as $P_{\rm SOTA}$. 

Secondly, we construct a one-hot probability vector from the ground truth human action for the current observation. We denote this probability as $P_{\rm gt}$.

Thirdly, we merge these two probabilities using hyperparameters 0.8 and 0.2 (we select the combination that yields the best result). We also zero out the actions that cause collisions. 

Finally, we renormalize the probability. The equation can be formulated as:
$$
P := {\rm Collision\,\,check\,\,and\,\,Renorm}(0.8P_{\rm SOTA} + 0.2 P_{\rm gt}).
$$

An example of the text output for $P$ is 

{\it ``Stop with probability 0.03, move forward with probability 0.55, turn left with probability 0.38, turn right with probability 0.00, look up with probability 0.03, and look down with probability 0.01''}

In this example, the action ``turn right'' has a probability of $0.00$ because we set it to zero due to a detected collision.

\section{Experiments}

We evaluate our method and compare it with baseline approaches on the Habitat-Matterport 3D Research Dataset (HM3D)~\cite{szot2021habitat}. We use the validation split from the HM3D-Semantics dataset~\cite{yadav2023habitat}, which consists of $20$ validation scenes. Following the evaluation pipeline of the work~\cite{ramrakhya2023pirlnav}, we report metrics on $2$k episodes. Our task is object goal navigation, where our agent starts at a random point within an indoor environment, and explores the environment until it reaches an object of a given object category (within $1$ meter distance). The exploration is limited to $500$ actions.

{\bf Metrics.} We report two metrics: success rate (Success) and soft success rate weighted by path length (SoftSPL). The Success measures the agent's ability to locate the target object goal within the allocated limit of permissible actions. Let $d_{\rm init}$ and $d_{\rm T}$ denote the geodesic distances to the target upon episode start and termination. The SoftSPL for an episode is defined as: ${\rm SoftSPL}=\Big( 1 - \frac{d_{\rm T}}{d_{\rm init}}\Big)\cdot \Big( \frac{s}{\max(s, p)}\Big)$, where $s$ and $p$ are the lengths of the shortest path and the path taken by the agent.

{\bf Baselines.} We compare four groups of baselines to evaluate our method. In the first group, we compare our method with non-behavior cloning methods. Specifically, we select two representative baselines: reinforcement learning (RL)~\cite{yadav2023habitat} and Goal-Oriented Semantic Exploration (SemExp)~\cite{chaplot2020object}. The RL baseline is trained using the DDPPO~\cite{wijmans2019dd} method without human demonstrations. On the other hand, SemExp constructs a top-down semantic map by combining the first-person semantic segmentation predictions with depth. It determines an exploration objective by considering the semantic map and the target object using a trained exploration policy. Furthermore, SemExp devises low-level actions to achieve this objective.

For the second group, we compare our method with state-of-the-art behavior cloning methods. We consider two baselines from the paper~\cite{ramrakhya2023pirlnav}: IL and RL\_Ft. IL~\cite{ramrakhya2023pirlnav} stands for imitation learning, which is learned purely based on behavior cloning. RL\_Ft~\cite{ramrakhya2023pirlnav} performs fine-tuning with reinforcement learning on top of the IL method. Note that, in Section~\ref{subsubsec:history_collector}, we pre-train our history collector model using the behavior cloning method. Hence, another way to understand our history collector is as the IL~\cite{ramrakhya2023pirlnav} method.

For the third group, we compare our method with three variants. The first variant involves using the pre-trained multimodal large language model (referred to as ${\rm LLM_{\rm no\,\,ft}}$) without fine-tuning to directly output the action for the agent based on the text prompt. We adopt MiniGPT4~\cite{zhu2023minigpt} for this variant. The second variant replaces the history collector model with $15$ consecutive observations (including images, GPS, compass values, and previous actions) to fine-tune LLMs, denoted as ${\rm LLM_{\rm consecutive\,\,obs}}$. In the third variant, instead of providing a probability output over actions, the model directly predicts the action itself, denoted as ${\rm LLM_{\rm direct\,\,action}}$.

For the fourth group, we aim to compare the impact of the collision check in our fine-tuning stage. In Section~\ref{subsec:finetune}, our method sets the probability to zero for the action that leads to a collision during training. In this case, we introduce a variant for the baseline that does not include a collision check, denoted as ${\rm LLM_{\rm no\,\,collision\,\,check}}$.

As a summary, we compare our method with RL, SemExp, IL, RL\_Ft, ${\rm LLM_{\rm no\,\,ft}}$, ${\rm LLM_{\rm consecutive\,\,obs}}$, ${\rm LLM_{\rm direct\,\,action}}$, and ${\rm LLM_{\rm no\,\,collision\,\,check}}$ approaches. IL and RL\_Ft are SOTA behavior cloning approaches, and the latter four are variants of our method.

\subsection{Comparisons with Non Behavior Cloning Methods}

Here, we compare methods with and without human demonstrations. Specifically, in Table~\ref{tbl:sota_bc}, we compare RL and SemExp with IL, RL\_Ft, and Ours. We observe that the methods trained without human demonstrations perform worse than the methods trained with human demonstrations. It is undeniable that human demonstrations provide us with exceptionally valuable information and can elevate the performance of models to the next level.

\subsection{Comparisons with SOTA Behavior Cloning Methods}

In this section, we compare our method with IL~\cite{ramrakhya2023pirlnav} and FL\_Ft~\cite{ramrakhya2023pirlnav} approaches. Both our method and the baselines utilize behavior cloning. The difference is that IL~\cite{ramrakhya2023pirlnav} and RL\_Ft~\cite{ramrakhya2023pirlnav} approaches are trained with non-large language models, while ours is fine-tuned using large language models. We present the results in Table~\ref{tbl:sota_bc}.

\begin{table}[t!]
\centering
\caption{Quantitative results for the comparisons among non-behavior-cloning methods, state-of-the-art behavior cloning methods, and our approach.}
\vspace{-2mm}
\begin{tabular}{c|cc}
\toprule
Methods        &       Success ($\uparrow$) & Soft SPL ($\uparrow$)   
\\
 \midrule \midrule
 \multicolumn{3}{c}{\it \small Non Behavior Cloning and No Large Language Models} \\
\midrule \midrule
RL~\cite{yadav2023habitat}   &    0.3936     & -                     \\ 
SemExp~\cite{chaplot2020object}   &   0.5560     &   -                       \\ 
 \midrule \midrule
 \multicolumn{3}{c}{\it \small Behavior Cloning without Large Language Models } \\
\midrule \midrule
 
IL~\cite{ramrakhya2023pirlnav}   &    0.5980     & 0.3051                     \\ 
RL\_Ft~\cite{ramrakhya2023pirlnav}   &   0.6615     &   0.3604                       \\ 
 \midrule \midrule
 \multicolumn{3}{c}{\it \small Behavior Cloning with Large Language Models } \\
\midrule \midrule
  Ours      &     {\bf 0.6790}     &    {\bf 0.3723}                    \\
       \bottomrule
\end{tabular}
\label{tbl:sota_bc}
\end{table}

First, we observe that RL\_Ft outperforms IL in terms of performance. It is important to note that IL is a pure imitation learning approach, while RL\_Ft is fine-tuned on top of IL using reinforcement learning. Therefore, we can conclude that reinforcement learning fine-tuning is beneficial. Second, we discover that our approach surpasses both IL and RL\_Ft, demonstrating the potential of LLMs to enhance visual navigation.

\subsection{Comparisons with LLMs Variants}

We present results that compare different variants of our approach using Language Models (LLMs). The results are shown in Table~\ref{tbl:variants}. The discussions in this section revolve around answering the following questions: ``{\it Does fine-tuning matter?}'', ``{\it Does the history collector help?}'', and ``{\it Direct action output or probability output?}''.

\begin{table}[t!]
\centering
\caption{Quantitative results for the comparisons among variants of our approach.}
\vspace{-2mm}
\begin{tabular}{c|cc}
\toprule
Methods        &       Success ($\uparrow$) & Soft SPL ($\uparrow$)   
\\
 \midrule \midrule
 \multicolumn{3}{c}{\it \small without Large Language Models Fine-tuning} \\
\midrule \midrule
 
${\rm LLM_{\rm no\,\,ft}}$   &    0.0000     & 0.0506                     \\ 
 \midrule \midrule
 \multicolumn{3}{c}{\it \small with Large Language Models Fine-tuning} \\
\midrule \midrule
 ${\rm LLM_{\rm consecutive\,\,obs}}$      &     0.0910     &    0.0977                    \\
  ${\rm LLM_{\rm direct\,\,action}}$     &     0.4610     &    0.2616                    \\
  Ours      &     {\bf 0.6790}     &    {\bf 0.3723}                    \\
       \bottomrule
\end{tabular}
\label{tbl:variants}
\end{table}

{\bf Does fine-tuning matter?} To address this question, we compare the results of ${\rm LLM_{\rm no\,\,ft}}$ with other approaches that involve fine-tuning of LLMs. We find that ${\rm LLM_{\rm no\,\,ft}}$ performs poorly, with a success rate of $0\%$. However, any method of fine-tuning can significantly improve visual navigation performance.

It is important to note that ${\rm LLM_{\rm no\,\,ft}}$ directly relies on a pre-trained large language model for action prediction, without using a history collector or performing fine-tuning. This approach is similar to zero-shot visual question answering experiments conducted in recent multimodal large language model research~\cite{zhu2023minigpt,li2023blip,dai2023instructblip,driess2023palm,huang2023instruct2act,brohan2023rt}. These studies reported success in those experiments. Therefore, the fact that zero-shot visual navigation in unfamiliar environments produces poor results indicates that visual navigation is a much more challenging problem compared to zero-shot visual question answering. Hence, fine-tuning is needed.

{\bf Does the history collector help?}
To answer this question, we compare ${\rm LLM_{\rm consecutive\,\,obs}}$ and ${\rm LLM_{\rm direct\,\,action}}$. The difference between these two approaches is that ${\rm LLM_{\rm consecutive\,\,obs}}$ considers input from 15 consecutive observations, while ${\rm LLM_{\rm direct\,\,action}}$ uses a history collector model to summarize all the information of the observations from the start of the episode until the current observation. We can clearly see a significant performance improvement from ${\rm LLM_{\rm consecutive\,\,obs}}$ to ${\rm LLM_{\rm direct\,\,action}}$, suggesting the benefits of using the history collector model.

\begin{figure}[t!]
    \centering    \includegraphics[width=0.95\linewidth]{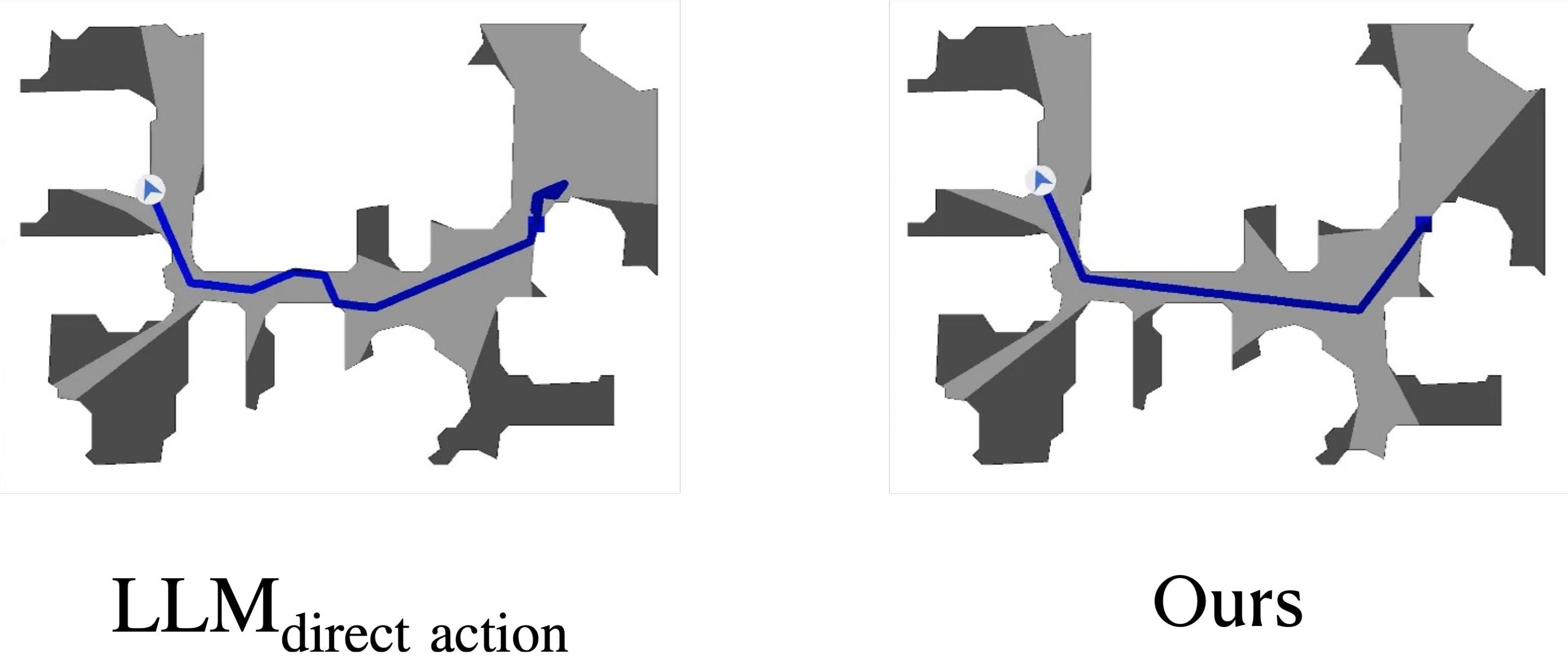}
    \vspace{-2mm}
    \caption{Qualitative results for comparing LLMs fine-tuned with visual navigation between direct action output and probability output. We show the results on the same scene, same initial location, and the same target object goal.}
    \label{fig:variants}
\end{figure}

{\bf Direct action output or probability output?} 
To answer this question, we compare ${\rm LLM_{\rm direct\,\,action}}$ and our approach. The main difference between these two approaches is that ${\rm LLM_{\rm direct\,\,action}}$ directly produces an action as output, whereas our approach outputs a probability distribution over all possible actions. Our results show that our approach significantly outperforms ${\rm LLM_{\rm direct\,\,action}}$, which suggests that, for visual navigation, generating probabilities as a form of uncertainty modeling is crucial. We also provide qualitative results comparing these two approaches in Figure~\ref{fig:variants}. The results show that our method has better path planning, with less turning and improved navigation in narrow aisles, compared to ${\rm LLM_{\rm direct\,\,action}}$.

\subsection{Comparisons on Collision Check}

Collision avoidance is crucial for visual navigation. In this section, we compare our method with a variant that does not include collision check (${\rm LLM_{no\,collision\,check}}$). In Section~\ref{subsec:finetune}, we specify that during training, we zero out the action that leads to a collision. ${\rm LLM_{no\,collision\,check}}$ simply removes this zeroing-out step. The results are reported in Table~\ref{tbl:collision}.

\begin{figure}[t!]
    \centering    \includegraphics[width=0.95\linewidth]{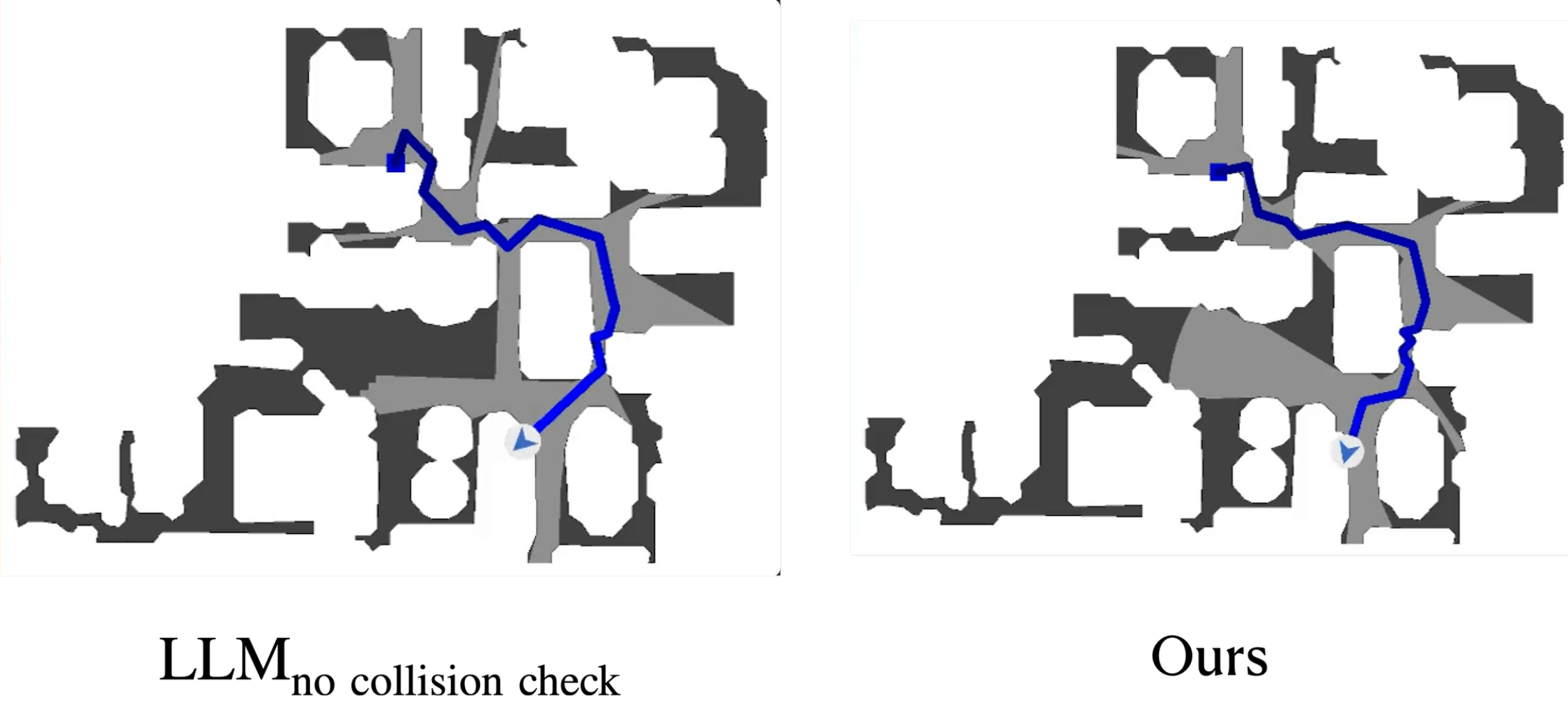}
    \vspace{-2mm}
    \caption{Qualitative results for comparing LLMs fine-tuned with visual navigation between with and without collision check. We show the results on the same scene, same initial location, and the same target object goal.}
    \label{fig:collision}
\end{figure}

Based on the numbers, it is evident that the collision check has a positive impact on all the metrics. It improves the success rate, enhances the SoftSPL, and reduces the collision count. These results indicate the significance of collision avoidance in visual navigation. Moving forward, our future work involves exploring and developing a more effective collision avoidance mechanism during the fine-tuning of LLMs. We also provide qualitative results comparing these two approaches in Figure~\ref{fig:collision}. The results indicate that the method can achieve fewer collisions when a collision check is performed.

\begin{table}[t!]
\centering
\caption{Quantitative results for the comparisons between our approach with and without collision check.}
\vspace{-2mm}
\begin{tabular}{c|cc}
\toprule
Methods        &   ${\rm LLM_{\rm no\,\,collision\,\,check}}$ & Ours  \\ 
\midrule

Success ($\uparrow$) & 0.6510 & {\bf 0.6790} \\
Soft SPL ($\uparrow$)  & 0.3641 & {\bf 0.3723} \\
Collision Count ($\downarrow$) & 39.7755 & {\bf 27.7615}     \\
       \bottomrule
\end{tabular}
\label{tbl:collision}
\end{table}

\section{Discussion and Future Work}

This work tackles the challenge of multimodal large language modeling with partial observation. In particular, the multimodal large language model is only able to access a limited portion of the overall environment and needs to navigate and explore the entire environment in order to complete tasks. In other words, our work involves addressing the setup of using multimodal large language models for long-horizon tasks. This approach is different from existing work on multimodal large language models, which usually assume full or nearly full observation. For example, previous work focuses on tasks like visual question answering given a specific image~\cite{zhu2023minigpt} or region grounding on a given image~\cite{yang2023dawn}. 

We argue that our work is the first to adopt multimodal large language models with partial observations, using visual navigation as a prime example. Other examples include long video generation, Atari game playing, and search and rescue operations. Due to the limited context length in large language models, it is not possible to directly feed all the information into the models. Therefore, a memory module is necessary to interact with the large language models. In the following, we present several potential solutions for adopting multimodal large language models with partial observations, using visual navigation as an example to illustrate these solutions.

{\bf Text-based RAG with Text Prompts.} One example of a memory module is extensive text-based prompt engineering system~\cite{zhou2023navgpt}. The large language models retrieve relevant contents from the extensive prompt engineering, incorporate them into the input, and generate a response based on the input. This process is known as augmented retrieval generation (RAG)~\cite{lewis2020retrieval}. While RAG has proven to be powerful for pure-text tasks, its performance with multimodal context has not been thoroughly studied. Representing all the multimodal context information directly in text may seem like an obvious solution, but prior work on visual navigation has shown that this approach is suboptimal~\cite{zhou2023navgpt}. 

{\bf Multimodal RAG with Multimodal Prompts.} In this approach, we create explicit memory with multimodal context. We retrieve relevant multimodal content from the memory and use it as a prompt for the large language models. For visual navigation, we can explore the idea of creating an SLAM (Simultaneous Localization And Mapping) system that generates maps in real-time. These maps can then be used as the relevant multimodal content. However, in order for the large language models to understand how to interpret maps, a fine-tuning process is necessary.

{\bf Multimodal Implicit Memory Module with Implicit Features Prompts.} In this approach, we do not focus on forming an explicit memory. Instead, we utilize neural networks as an implicit memory module to condense all past information into a fixed-dimensional feature. This implicit feature is then fed into the large language models. The advantage of this approach is its simplicity, as there is no need to select the most appropriate multimodal content for the language models. However, the effectiveness of this approach heavily relies on the design and training of the implicit memory module (neural networks). Our paper follows this approach, using GRUs as the implicit memory module and training them with the same dataset as the large language models. Lastly, similar to the multimodal RAG with multimodal prompts, fine-tuning of the large language models is necessary to understand the implicit feature as a prompt.

\subsection{More on Data}

In this paper, we propose using human demonstrations as the training data for multimodal large language models with partial observations. Human demonstrations have the advantage of being high quality and low noise. However, collecting human demonstrations can be expensive, so it is important to consider other data sources as well.

In the context of visual navigation, prior work~\cite{ramrakhya2023pirlnav} also explores using {\it shortest path} and {\it frontier exploration} as data sources. These data are easier to collect since they can be automatically gathered without human intervention. However, the quality of these data is not guaranteed, resulting in models trained with these data performing less favorably compared to models trained with human demonstrations. Taking inspiration from Tesla's data collection efforts, we argue the best approach is to curate human demonstrations with extensive data augmentations from simulation. 

\subsection{More on Training}

The concept of learning with partial observations or learning with long horizon tasks is often discussed in the reinforcement learning (RL) literature~\cite{sutton2018reinforcement}. Therefore, in addition to the behavior cloning algorithm, RL algorithms can be a potential alternative for training or fine-tuning large language models. However, there is currently no evidence to suggest that RL algorithms can effectively work with large language models.

The main challenge for RL in training large language models is the sparse nature of the supervision signals. We argue that training large language models requires strong, dense, and semantically meaningful supervision signals. We demonstrate an example of dense and semantically meaningful supervision signals in our paper, where the output for the large language models is designed to be a probability distribution over all possible actions.

To enable RL training with large language models, we need to convert sparse supervision signals into dense and semantically meaningful signals. However, this problem remains unsolved in the RL community~\cite{kaelbling1996reinforcement}. One potential workaround is to consider unsupervised auxiliary tasks~\cite{jaderberg2016reinforcement}, such as predicting the next action or predicting the next input. In summary, we believe that RL can be a potential method for training and fine-tuning multimodal large language models with partial observations. However, there is still a long way to go, and significant efforts are required to address the challenges.

\subsection{What's the next step?}

So far, we have discussed several solutions, data, and training methods for adopting multimodal large language models with partial observations. For our next step, we plan to investigate the following: 1) multimodal RAG with multimodal prompts, 2) data augmentations on human demonstration data using simulators, and 3) exploring other datasets or tasks.

\section{Conclusion}

In this paper, we explore the fine-tuning of Large Language Models (LLMs) for visual navigation. Unlike previous work, which focuses on complex prompt engineering for visual navigation using LLMs, our approach is simple. We use a basic text prompt, a history collector model that incorporates tokens from past observations, an observation encoding model that embeds observation tokens, and a pre-trained large language model. During training, we employ two tricks based on human demonstrations. First, instead of directly outputting the action for the agent, we output the probability distribution over all possible actions. Second, we construct this probability distribution using a state-of-the-art behavior cloning method, the action demonstrated by a human, while avoiding actions that cause collisions. We believe that our work highlights the advantages of fine-tuning LLMs for visual navigation. Our experimental results support this claim, as our approach outperforms state-of-the-art methods.

{\small
\bibliographystyle{ieee_fullname}
\bibliography{ref}
}


\end{document}